\documentclass{article}

\usepackage{times}
\usepackage{graphicx} % more modern
\usepackage{epsfig} % less modern
\usepackage{subfigure} 

\usepackage{natbib}

\usepackage{algorithm}
\usepackage{algorithmic}

\usepackage{amsmath}
\usepackage{amssymb}

\usepackage{hyperref}

\newtheorem{theorem}{Theorem}
\newtheorem{corollary}{Corollary}

\usepackage[accepted]{icml2015}

\icmltitlerunning{Perfect Machine Learning}

\begin{document} 

\twocolumn[
\icmltitle{
A New Perspective on Machine Learning: \\
How to do Perfect Supervised Learning}

% unless you've provided the [accepted] option to the icml2015
\icmlauthor{Hui Jiang}{hj@cse.yorku.ca}
\icmladdress{
iFLYTEK Laboratory for Neural Computing and Machine Learning (iNCML) \\
Department of Electrical Engineering and Computer Science  \\
York University,  4700 Keele Street, Toronto, Ontario, M3J 1P3, Canada \\
\url{http://wiki.cse.yorku.ca/user/hj/}}

\icmlkeywords{machine learning, supervised learning, bandlimited, langrage interpolation}

\vskip 0.3in
]

\begin{abstract} 
In this work, we introduce the concept of bandlimiting into the theory of machine learning because all physical processes are bandlimited by nature, including real-world machine learning tasks. After the bandlimiting constraint is taken into account, our theoretical analysis has shown that all practical machine learning tasks are asymptotically solvable in a perfect sense. Furthermore, the key towards this solvability almost solely relies on two factors: i) a sufficiently large amount of training samples beyond a threshold determined by a difficulty measurement of the underlying task; ii) a sufficiently complex and bandlimited model.  Moreover, for some special cases, we have derived new error bounds for perfect learning, which can quantify the difficulty of learning. These generalization bounds are not only asymptotically convergent but also irrelevant to model complexity. Our new results on generalization have provided a new perspective to explain the recent successes of large-scale supervised learning using complex models like neural networks.
\end{abstract} 

\section{Introduction}
\label{sec_intro}

The fundamental principles and theories for machine learning (ML) were established a few decades ago, such as the No-Free-Lunch theory \cite{Wolpert95}, statistical learning theory \cite{Vapnik00}, and probably approximately correct (PAC) learning \cite{Valiant84}. These theoretical works have successfully explained which problems are learnable and how to achieve effective learning in principle. On the other hand, since the boom of deep learning in the past decade, the landscape of machine learning practices has changed dramatically. A variety of artificial neural networks (ANN) have been successfully applied to all sorts of real-world applications, ranging from speech recognition and image classification to machine translation. The list of success stories in many diverse application domains is still growing year after year. The superhuman performance has even been claimed in some tasks, which were originally thought to be very hard. The divergence between the theories and the practices has equally puzzled both ML theorists and ML practitioners. At this point, we desperately need to answer a series of serious questions in order to further advance the field as a whole. For instance, why do the ANN-type models significantly overtake other existing ML methods on all of these practical applications? What is the essence to the success of the ANN-type models on these ostensibly challenging tasks? Where is the limit of these ANN-type models? Why does horrific overfitting, as predicted by the current ML theory, never happen in these real-world tasks even when some shockingly huge models are used? \cite{Zhang16} 

In this paper, we develop a new ML theory to shed some light on these questions. The key to our new theory is the concept of bandlimiting. Not all processes may actually exist in the real world and all physically realizable processes must be bandlimited. Much of the previous efforts in machine learning theory have been spent in studying some extremely difficult problems that are over-generalized in theory but may not actually exist in practice. After the bandlimiting constraint is taken into account, our theoretical analysis has shown that all practical machine learning tasks are asymptotically solvable in a perfect sense. Our theoretical results suggest that the roadmap towards successful supervised learning consists of several steps: (a) collecting sufficient labelled in-domain data; (b) fitting a complex and {\it bandlimited} model to the large training set. The amount of data needed for perfect learning depends on the difficulty of each underlying task. For some special cases, we have derived some new error bounds to quantitatively measure the difficulty of learning. As the amount of training data grows, we need a complicated model to complement step (b). The universal approximation theory in \cite{Cybenko89,Hornik91} makes neural networks an ideal candidate for perfect learning since similar model structures can be fitted to any large training set if we keep increasing the model size. The highly-criticized engineering tricks used in the training of neural networks are just some empirical approaches to ensure that a complicated model is effectively fit to a very large training set in step (b) \cite{Jiang19b}. However, there is no evidence to show that neural networks are the only possible models that are able to achieve perfect learning. 

\section{Problem Formulation}
\label{sec_formulation}

In this work, we study the standard supervised learning problem in machine learning. Given a finite training set of $N$ samples of input and output pairs, denoted as
$\mathcal{D}_N = \{ ( \mathbf{x}_1, y_1 ),  ( \mathbf{x}_2, y_2 ), \cdots, ( \mathbf{x}_N, y_N )\}
$, the goal is to learn a model from input to output over the entire feature space: $\mathbf{x} \to y$, which will be used to predict future inputs.  

\subsection{Machine Learning as Stochastic Function Fitting}

Instead of starting our analysis from the joint probabilistic distribution of inputs and outputs $p(\mathbf{x}, y)$ as in normal statistical learning theory, we adopt a more restricted formulation in this paper. Here, we assume all inputs $\mathbf{x}$ are random variables following a probabilistic density function, i.e. $\mathbf{x} \sim p(\mathbf{x})$, in the input feature space (without losing generality, we may assume $\mathbf{x} \in \mathbb{R}^K$). The relation between input $\mathbf{x}$ ($\mathbf{x} \in \mathbb{R}^K$) and output $y$ ($y \in \mathbb{R} $) is deterministic, which may be represented by a function $f: \mathbf{x} \to y \;\; ( \mathbf{x} \in \mathbb{R}^K, y \in \mathbb{R})$, denoted as the target function $y=f(\mathbf{x})$. In this setting, the goal of machine learning is to learn a model $\hat{f}(\mathbf{x})$ to minimize the expected error between $\hat{f}(\mathbf{x})$ and $f(\mathbf{x})$ as measured by $p(\mathbf{x})$.

Most of interesting and meaningful learning problems in the real world can be easily accommodated by the above deterministic function between inputs and outputs.  For example, we may define $y=f(\mathbf{x}) = \arg\max_{y} p(y|\mathbf{x})$ if the conditional distribution $p(y|\mathbf{x})$ is sharp and unimodal. If $p(y|\mathbf{x})$ is sharp but not unimodal, we may decompose the learning problem into several sub-problems, each of which is represented by one deterministic function as above. If $p(y|\mathbf{x})$  is not sharp, it means that the relation between inputs and outputs are fairly weak. In these cases, either it is usually not a meaningful learning problem in practice, or we may improve input features $\mathbf{x}$ to further enhance the relation between  $\mathbf{x}$ and $y$. 

\subsection{The Bandlimiting Property}

\begin{figure}[h!]
\begin{subfigure}%{\textwidth}
  \centering
  \caption{A function without band-limiting }
  \includegraphics[width=1\linewidth]{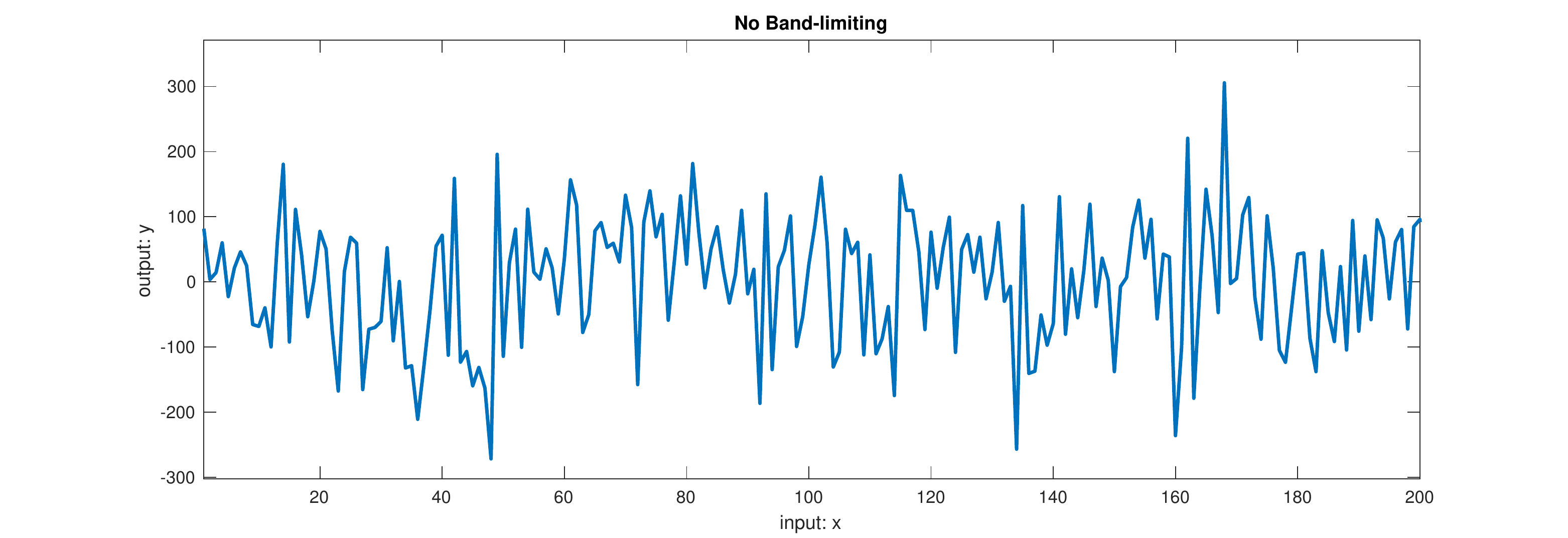}
\label{Fig-bandlimited-functions-fig1}
\end{subfigure}%
\begin{subfigure}%{\textwidth}
  \centering
  \caption{A weakly band-limited function  by a large limit}
  \includegraphics[width=1\linewidth]{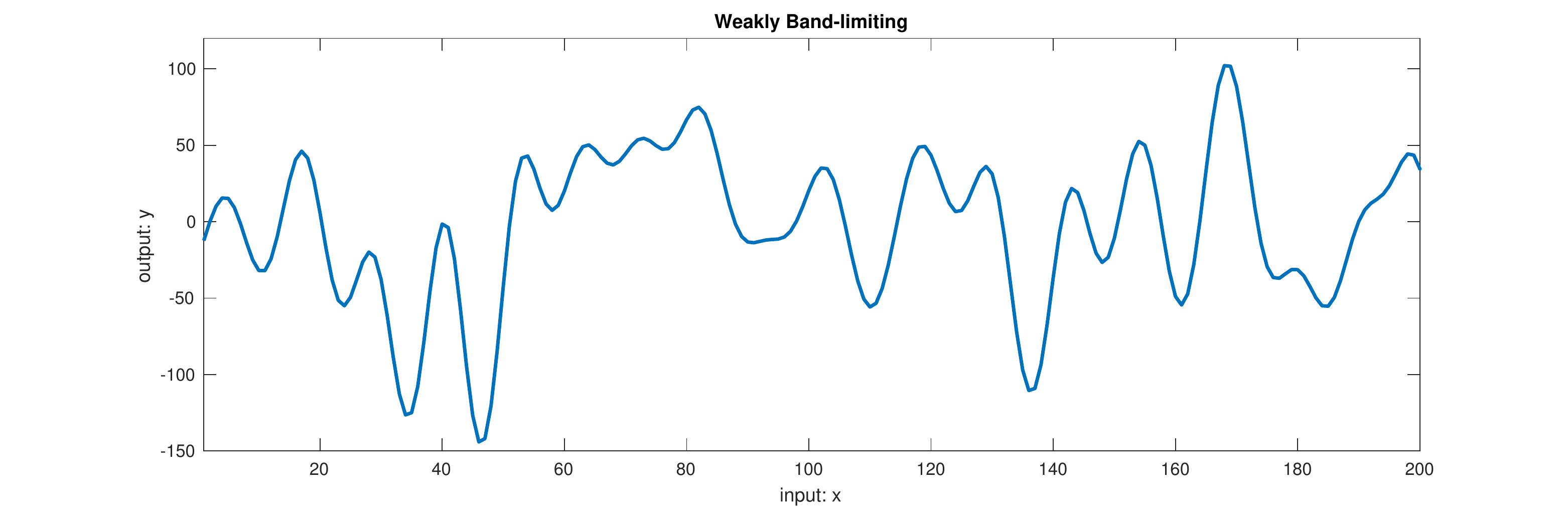}
\label{Fig-bandlimited-functions-fig2}
\end{subfigure}
\begin{subfigure}%{.5\textwidth}
  \centering
  \caption{A strongly band-limited function by a small limit}
  \includegraphics[width=1\linewidth]{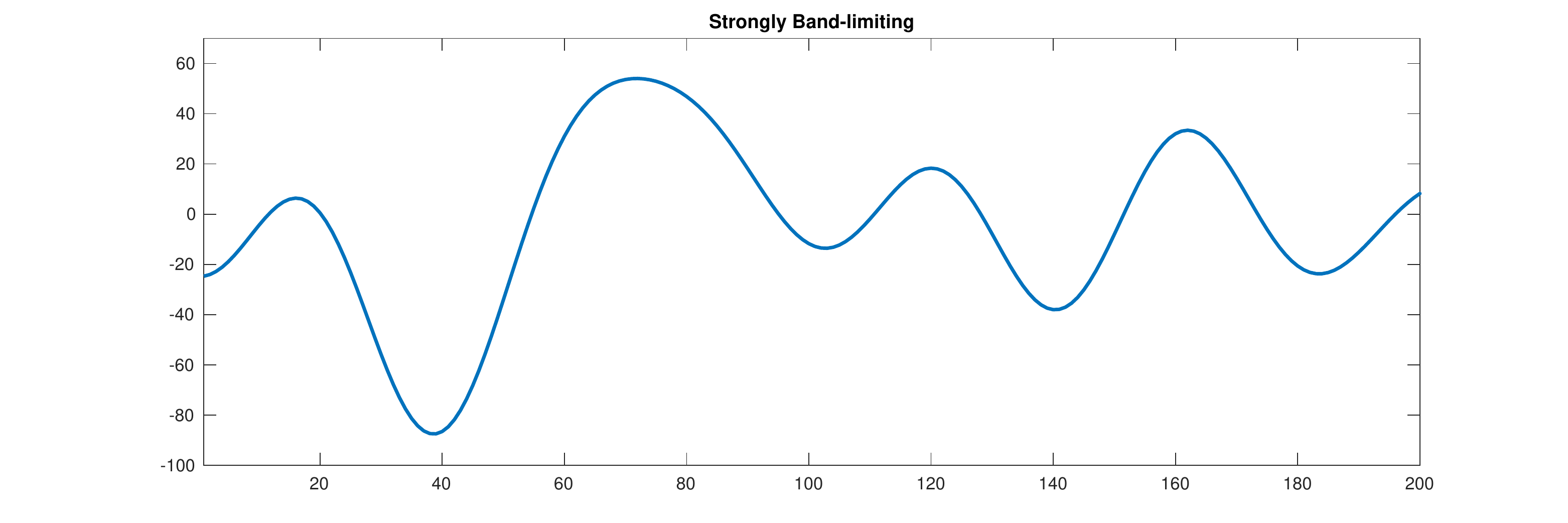}
\label{Fig-bandlimited-functions-fig3}
\end{subfigure}
\begin{subfigure}%{.5\textwidth}
  \centering
  \caption{An approximately band-limited function}
  \includegraphics[width=1\linewidth]{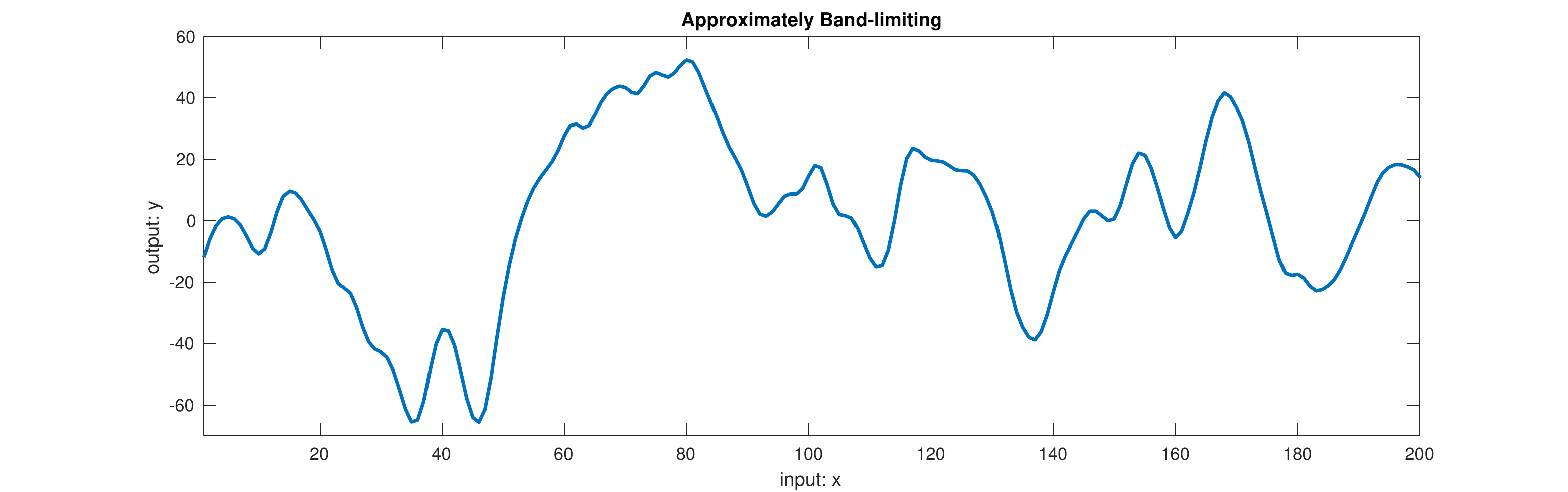}
\label{Fig-bandlimited-functions-fig4}
\end{subfigure}
%\caption{An illustration to show the property of various types of 1-D bandlimiting functions}
\end{figure}

In engineering, it is a well-known fact that all physically realizable processes must satisfy the so-called {\it bandlimiting} property. Bandlimiting is a strong constraint imposed on the smoothness and growth of functions,
which corresponds to the mathematic concept of finite exponent type of entire functions in mathematical analysis \cite{Levin64,Levinson40}. 
As shown in Figures \ref{Fig-bandlimited-functions-fig1} to \ref{Fig-bandlimited-functions-fig4}, 
several 1-D functions with different bandlimiting constraints are plotted as an illustration, which clearly show that the various bandlimiting contraints heavily affect the smoothness of a function. 

In practice, if a supervised learning problem arises from a real-world task or a physical process, the above target function $y=f(\mathbf{x})$ will satisfy the bandlimiting property as constrained by the physical world. The central idea in this paper is to demonstrate that the bandlimiting property, largely overlooked by the machine learning community in the past, is essential in explaining why real-world machine learning problems are not as hard as speculated by statistical learning theory \cite{Vapnik00,Shai14}. The theory proposed in this paper further suggests that under certain conditions we may even solve many real-world supervised learning problems perfectly. 

First of all, let's give the definition of {\it bandlimiting}\footnote{Also known as {\it wavenumber-limited}. Here, we prefer the term ``bandlimiting'' as it is better known in engineering. }. A function
$f(\mathbf{x}) \in L^1(\mathbb{R}^K)$ %(or $f(\mathbf{x})p(\mathbf{x}) \in L^1$ if necessary) 
is called to be strictly bandlimited if its multivariate Fourier transform \cite{Stein71}, $F(\boldsymbol{\omega})$, vanishes to zero beyond a certain finite spatial frequency range. If there exists $B>0$, such that
\begin{equation}\label{eq-strict-bandlimiting}
F(\boldsymbol{\omega}) = 
\idotsint_{-\infty}^{+\infty} 
f(\mathbf{x})  e ^{-i \mathbf{x} \cdot \boldsymbol{\omega} } \; d\mathbf{x}
= 0  \;\; \mbox{if} \;\; \Vert \boldsymbol{\omega} \Vert > B,
\end{equation} 
then $f(\mathbf{x})$ is called a strictly bandlimited function by $B$.

Similarly, we may define a function $f(\mathbf{x}) \in L^1(\mathbb{R}^K)$  is approximately bandlimited if 
its Fourier transform $F(\boldsymbol{\omega})$ satisfies:
$$ %\begin{equation} 
\lim_{B \to +\infty} 
\idotsint_{\Vert{\omega}\Vert > B} 
\Vert F(\boldsymbol{\omega}) \Vert^2  \; d\mathbf{\boldsymbol{\omega}}
= 0. 
$$ %\end{equation}

In other words, for any arbitrary small $\epsilon >0$, $\exists B_\epsilon >0$, the out-of-band residual energy satisfies
\begin{equation}\label{eq-approximate-bandlimiting2}
\idotsint_{\Vert{\omega}\Vert > B_\epsilon} 
\Vert F(\boldsymbol{\omega}) \Vert^2  \; d\mathbf{\boldsymbol{\omega}}
< \epsilon^2
\end{equation}
where $B_\epsilon$ is called the approximate band of $f(\mathbf{x})$ at $\epsilon$.

\section{Perfect Learning}
\label{sec_mainresults}

In supervised learning, we are interested in learning the unknown target function $y=f(\mathbf{x})$ based on a finite training set of $N$ samples of input and output pairs:
$\mathcal{D}_N = \Big\{ ( \mathbf{x}_1, y_1 ),  ( \mathbf{x}_2, y_2 ), \cdots, ( \mathbf{x}_N, y_N ) \Big\},
$
where each pair $(\mathbf{x}_i, y_i)$  ($1 \leq i \leq N$ ) is an i.i.d. sample and  $\mathbf{x}_i$ is randomly drawn from an unknown p.d.f $p(\mathbf{x})$, i.e. $\mathbf{x}_i \sim p(\mathbf{x})$ and $y_i = f(\mathbf{x}_i)$.
The central issue in supervised learning is how to learn a model from the given training set $\mathcal{D}_N$, denoted as $ \hat{f}(\mathbf{x} | \mathcal{D}_N)$, in order to minimize the so-called {\it expected risk}, $R(\hat{f})$, defined over the entire feature space in the sense of mean squared error (MSE):
\begin{eqnarray}
R(\hat{f}\, |\,  \mathcal{D}_N)  & = & \mathbf{E}_{p(\mathbf{x})} \Big[ \Vert\hat{f}(\mathbf{x} | \mathcal{D}_N) - f(\mathbf{x}) \Vert ^2  \Big]  \nonumber \\
& \hspace{-4cm} = & \hspace{-2cm} \idotsint_{-\infty}^{+\infty}
\left\Vert f(\mathbf{x}) - \hat{f}(\mathbf{x} | \mathcal{D}_N) \right\Vert^2 
p(\mathbf{x}) \; d\mathbf{x}.
\end{eqnarray}

Usually the above expected risk is not practically achievable since it requires two unknown functions, $f(\mathbf{x})$ and $p(\mathbf{x})$. Supervised learning methods instead focus on learning a model $\hat{f}$ to optimize the so-called {\it empirical risk}, computed solely on the given training samples as follows:
\begin{equation}
R_{emp} (\hat{f} \, |\,  \mathcal{D}_N)  = \frac{1}{N}
\sum_{i=1}^N \;\; \left\Vert y_i - \hat{f} (\mathbf{x}_i  | \mathcal{D}_N) \right\Vert^2
\end{equation}

Here we use MSE for mathematic simplicity but our analysis is equally applicable to both regression and classification problems. 
We know that the unknown expected risk is linked to the above empirical risk by uniform bounds in the VC theory \cite{Vapnik00} for classification. In machine learning, it is common practice to apply some sort of regularization to ensure these two quantities will not diverge in the learning process to avoid the so-called overfitting. 

\subsection{Existence of Perfect Learning}

In this work, we define {\it perfect supervised learning} as an ideal scenario where we can always learn a model from a finite set of training samples as above to achieve not only zero empirical risk but also {\rm zero expected risk.} Here, we will theoretically prove that perfect supervised learning is actually achievable if the underlying target function is bandlimited and the training set is sufficiently large. 

\begin{theorem} (existence)
\label{theorem-exisitence}
In the above supervised learning setting, if the target function $f(\mathbf{x})$ is strictly or approximately bandlimited, given a sufficiently large training set ${\cal D}_N$ as above, then there exists a method to learn a model (or construct a function) $\hat{f}(\mathbf{x}|{\cal D}_N)$ solely from ${\cal D}_N$, not only leading to zero empirical risk
$$
R_{emp} (\hat{f} \, |\,  \mathcal{D}_N) = 0
$$
but also yielding  zero expected risk in probability 
$$
R(\hat{f} \, |\,  \mathcal{D}_N)  \overset{P}{\longrightarrow} 0 %\;\;\;\mbox{as} \;\;\; N \to \infty.
$$
as $N \to \infty$.
\end{theorem}

{\bf Proof sketch:} The idea is similar to the multidimensional sampling theorem \cite{Petersen62}, stating that a bandlimited signal may be fully represented by infinite uniform or non-uniform samples as long as these samples are dense enough \cite{Marvasti01}. In our case, we attempt to recover the function $f(\mathbf{x})$ from the samples randomly drawn according to a probability distribution. Obviously, as $N \to \infty$, they will surely satisfy any density requirement determined by the band of $f(\mathbf{x})$. Moreover, we will show that the truncation error from infinite samples to $N$ finite samples is negligible and will vanish when $N \to \infty$.
See the full proof in Appendix \ref{appendix-proof-theorem1}. 
$\hspace*{\fill} {\hfill\blacksquare}$

This result may theoretically explain many recent successful stories in machine learning. As long as a learning task arises from any real-world application, no matter whether it is related to speech, vision, language or others, it is surely bounded by the bandlimitedness property in the physical world. As long as we are able to collect enough samples, these problems will be solved almost perfectly by simply fitting a complex model to these samples in a good way. The primary reason for these successes may be attributed to the fact that these real-world learning problems are not as hard as they were initially thought to be. At a glance, these problems are regarded to be extremely challenging due to the involved dimensionality and complexity. However, the underlying processes may in fact be heavily bandlimited by some incredibly small values. 

On the other hand, it is impossible to achieve perfect learning if the target function $f(\mathbf{x})$ is not bandlimited. 
\begin{corollary} \label{corollary-nonexisitence}
If $f(\mathbf{x})\cdot p(\mathbf{x})$ is not strictly nor approximately bandlimited, no matter how many training samples to use, $R(\hat{f}\, |\,  \mathcal{D}_N)$ of all realizable learning algorithms have a nonzero lower-bound:
$$
\lim_{N \to \infty} \; R(\hat{f} \, |\,  \mathcal{D}_N)  \; \geq \varepsilon  \;> 0.
$$
\end{corollary}

\subsection{Non-asymptotic Analysis}

The previous section gave some results on the asymptotic behaviour of perfect supervised learning when $N \to \infty$. Here, let us consider some non-asymptotic analyses to indicate how hard a learning problem may be when $N$ is finite. 
Given any one training set of $N$ i.i.d. samples $\mathcal{D}_N$,
we may learn a model, denoted as $\hat{f}(\mathbf{x} | {\cal D}_N)$ 
from $\mathcal{D}_N$. If $N$ is finite, when we select different training sets of $N$ samples, the same learning algorithm may end up with a different result each time. In this case,  the learning performance should be measured by the {\it mean expected risk} averaged with respect to $\mathcal{D}_N$:
\begin{equation} \label{eq-average-expected-risk}
{\cal R}_N = {\bf E}_{ \mathcal{D}_N } \left[   \mathbf{E}_{p(\mathbf{x})} \left[ \Vert\hat{f}(\mathbf{x} | \mathcal{D}_N) - f(\mathbf{x}) \Vert ^2  \right]     \right]
\end{equation}

\subsubsection{Strictly Bandlimited Target Functions}

For any finite $N$ and strictly bandlimited target functions $f(\mathbf{x})$,  
we first consider a simple case, where $\mathbf{x}$ follows an isotropic covariance Gaussian distribution. 
We  have the following result to upper-bound the above mean expected risk in eq. (\ref{eq-average-expected-risk}) for the perfect learning algorithm:

\begin{theorem} 
\label{theorem-error-bound-gaussian}
If we have $\mathbf{x} \sim p(\mathbf{x}) = {\cal N}({\bf 0}, \sigma^2 {\bf I})$, the target function $f(\mathbf{x})$ is strictly bandlimited by $B$, 
the mean expected risk in eq.(\ref{eq-average-expected-risk}) of the perfect learner 
is upper bounded as follows:
\begin{equation}\label{eq-HJ-bound-strictband}
{\cal R}^*_N  < \left[ \frac{(\sqrt{2} K B\sigma)^{n+1} \cdot H} {\sqrt{(n+1)! }}
 \right] ^2 
\end{equation}
where $K$ is the dimension of $\mathbf{x}$, $n \simeq O(N^{1/K})$ and $H=\sup_{\mathbf{x}} |f(\mathbf{x})|$ is the maximum value of $f(\mathbf{x})$.
\end{theorem}

{\bf Proof sketch:} Based on the given $N$ samples, assume a model is learned as multivariate Taylor polynomials of $f(\mathbf{x})$ up to certain order $n$, which exactly has $N$ free coefficients. This error bound may be derived based on the remainder error in the multivariate Taylor's theorem.
See  the full proof in Appendix \ref{appendix_derivatives_eq}.
$\hspace*{\fill} {\hfill\blacksquare}$

This bound in Theorem \ref{theorem-error-bound-gaussian} serves as a general indicator for how hard a learning problem is. It also suggests that learning is fairly easy when the target function is bandlimited by a finite range $B$, where the mean expected risk of a good learning algorithm may converge {\em exponentially} to 0 as $n \to \infty$ (when $N \to \infty$). When $N$ is relatively small, the difficulty of the learning problem is well-reflected by the quantity of $K B \sigma$. 
$K$ is the dimensionality of the underlying problems: it is not necessarily equal to the dimensionality of the raw data since those dimensions are highly correlated, and it may represent the dimensionality of the independent features in a much lower de-correlated space. Note that $K$ also affects the convergence rate of learning since $n \simeq O(N^{1/K})$. Generally speaking, the larger the value $K B \sigma$ is, the more difficult the learning task will be and the more training samples are needed to achieve good performance. 
For the same number of samples from the same data distribution $p(\mathbf{x})$, it is easier to learn a narrowly-banded function than a widely-banded one. On the other hand, in order to learn the same target function $f(\mathbf{x})$ using the same number of samples, it is much easier to learn in the cases where the data distribution is heavily concentrated in the space than those where the data is wildly scattered.

Moreover, we can easily extend Theorem  \ref{theorem-error-bound-gaussian} to diagonal covariance matrices. 
\begin{corollary} \label{corollary-diagonalcovariance-bound}
If $\mathbf{x}$ follows a multivariate Gaussian distribution with zero mean and diagonal covariance matrix, ${\cal N}(0, \Sigma)$, with $\Sigma = \mbox{diag} [ \sigma^2_1, \sigma^2_2, \cdots, \sigma^2_K]$, and the target function $f(\mathbf{x})$ is bandlimited by different values $B_k$ ($k=1,2,\cdots,K$) for various dimensions of $\mathbf{x}$, we have  
\begin{equation}\label{eq-HJ-bound-strictband-diagonal}
{\cal R}^*_N  <
\frac{1}{K}\sum_{k=1}^K \; \left[ \frac{(\sqrt{2} K B_k\sigma_k)^{n+1} \cdot H} {\sqrt{(n+1)! }}
 \right] ^2 
\end{equation}
\end{corollary}
In this case, different dimensions may contribute to the difficulty of learning in a different way. In some high dimensional problems, many dimensions may not affect the learning too much if the values of  $B_k \sigma_k$ are negligible in these dimensions. 

At last, we give a fairly general case for strictly bandlimited functions $f(\mathbf{x})$. Assume $\mathbf{x} \sim p(\mathbf{x})$ is constrained in a bounded region in $\mathbb{R}^K$, we may normalize all $\mathbf{x}$ within a hypercube, denoted as $[-U, U]^K$.

\begin{corollary} 
If $\mathbf{x}$ follows any distribution $p(\mathbf{x})$ within a hypercube $[-U, U]^K \subset \mathbb{R}^K$, and the target function $f(\mathbf{x})$ is bandlimited by $B$, the perfect learner is upper-bounded as:
\begin{equation}
{\cal R}^*_N  < \left[ \frac{(K B U)^{n+1} \cdot H} {(n+1)!}
 \right] ^2
\end{equation}
\end{corollary}

\subsubsection{Approximately Bandlimited Target Functions} 

Assume the target function $y=f(\mathbf{x})$ is not strictly bandlimited by any fixed value $B$, but approximately bandlimited as in eq.(\ref{eq-approximate-bandlimiting2}).  Here, we consider how to compute the expected error for a given training set of $N$ samples, i.e., $\mathcal{D}_N$.
In this case, for any arbitrarily small $\epsilon >0$, we may have an approximate band $B_\epsilon$ to decompose the original function $y=f(\mathbf{x})$ into two parts: $f(\mathbf{x}) = f_{B}(\mathbf{x}) + f_e(\mathbf{x})$, where $f_{B}(\mathbf{x})$ is strictly bandlimited by $B_\epsilon$ and $f_e(\mathbf{x})$ contains the residual out of the band. As shown in eq.(\ref{eq-approximate-bandlimiting2}), we have 
$ \idotsint_{-\infty}^{\infty}
\Vert F_e(\boldsymbol{\omega}) \Vert^2  \; d\mathbf{\boldsymbol{\omega}}
< \epsilon^2
$, where $F_e(\boldsymbol{\omega})$ is the Fourier transform of the residual function $f_e(\mathbf{x})$.

If $\mathbf{x} \sim {\cal N}({\bf 0}, \sigma^2 {\bf I})$,  following Theorem \ref{theorem-error-bound-gaussian} and Parseval's identity, we have
$$
{\cal R}^*_N  <  \left[ \frac{(\sqrt{2} K B_{\epsilon}\sigma)^{n+1} \cdot H} {\sqrt{(n+1)! }}
 \right] ^2 + \epsilon^2 
$$
where the second term is the so-called aliasing error. For any given problem setting, if we decrease $\epsilon$, the first term becomes larger since $B_\epsilon$ is larger. Therefore, we can always vary $\epsilon$ to look for the optimal $\epsilon_*$ to further tighten the bound on the right hand side of the above equation as:
$
{\cal R}^*_N  \leq  \left[ \frac{(\sqrt{2} K B_{\epsilon_*}\sigma)^{n+1} \cdot H} {\sqrt{(n+1)! }}
 \right] ^2 + \epsilon_*^2.
$

\section{Conditions of Perfect Learning}

Here we study under what conditions we may achieve the perfect learning in practice. First of all, the target function must be bandlimited, i.e., all training data are generated from a bandlimited process. Secondly, when we learn 
a model from a class of strictly or approximately bandlimited functions, if the learned model achieves the zero empirical risk on a sufficiently large training set, 
then the learned model is guaranteed to yield zero expected risk for sure.
In other words, under the condition of bandlimitedness, the learned model will naturally generalize to the entire space if it fits to a sufficiently larget training set. 

\begin{theorem}  (sufficient condition)
\label{theorem-condition}
If the target function $f(\mathbf{x})$ is strictly or approximately 
bandlimited, 
assume a strictly or approximately bandlimited model, $\hat{f}(\mathbf{x})$, is learned from a sufficiently large training set ${\cal D}_N$.
If this model yields zero empirical risk on ${\cal D}_N$:
$$
R_{emp} (\hat{f} \, |\,  \mathcal{D}_N) = 0,
$$
then it is guaranteed to yield zero expected risk:
$$
R(\hat{f}\, |\,  \mathcal{D}_N)  \longrightarrow 0
$$
as $N \to \infty$.
\end{theorem}
{\bf Proof sketch:}  If $f(\mathbf{x})$ and  $\hat{f}(\mathbf{x})$ are bandlimited, each of them may be represented as an infinite sum of diminishing terms. If a bandlimited model $\hat{f}(\mathbf{x})$ is fit to a bandlimited  target function $f(\mathbf{x})$ based on $N$ training samples, it ensures that the $N$ most significant terms of $\hat{f}(\mathbf{x})$ are learned up to a good precision. As $N \to \infty$, the learned model $\hat{f}(\mathbf{x})$ will surely converge to the target function $f(\mathbf{x})$.
 See the full proof in Appendix \ref{appendix-proof-condition}. 
$\hspace*{\fill} {\hfill\blacksquare}$

This theorem gives a fairly strong condition for generalization in practical machine learning scenerios. In practice, all real data are generated from a bandlimited target function. If we use a bandlimited model to fit to a large enough training set, the generalization of the learned  model is guaranteed asymptotically by itself. Under some minor conditions, namely the input and all model parameters are bounded, it is easy to show that all continuous models are at least approximately bandlimited, including most PAC-learnable models widely used in machine learning, such as linear models, neural networks, etc. In these cases, perfect learning mostly rely on whether we can perfectly fit the model to the given large training set. In our analysis, model complexity is viewed as an essence towards the success of learning because complex models are usually needed to fit to a large training set. Our theorems show that model complexity does not impair the capability to learn as long as the complex models satisfy the bandlimitedness requirement. 
Bandlimitedness is a model characteristic  orthogonal to model complexity (which is reflected by the number of free parameters). We may have a simple model that has an unlimited spatial frequency band. On the other hand, it is possible to have a very complex model which is strongly bandlimited by a small value
\footnote{See more explanation in paragraph 4 of Appendix \ref{appendix_derivatives_eq}}.
On the other hand, the traditional statistical learning theory leads to fairly loose bounds for simple models and completely fails to explain complex models due to the huge or even infinite VC dimensions. 

This theorem will help to explain the generalization magic of neural networks recently observed in the deep learning community \cite{Zhang16}. 
As discussed above, when the input and all model parameters of any neural network are bounded, we may normalize the input into a hypercube $[-U,U]^K$, in this case, the function represented by a neural network belong to the function class $L^1([-U,U]^K)$. According to the Riemann-Lebesgue lemma \cite{Pinsky02}, the Fourier transform of any function in $L^1([-U,U]^K)$ decays when the absolute value of any frequency component goes up. Therefore, any neural network is essentially an approximately bandlimited model. Based on the Theorem \ref{theorem-condition}, we can easily derive the following corollary.

\begin{corollary} \label{corollary-NN-generalization}
Assume a neural network, $\tilde{f}(\mathbf{x})$, is learned from a sufficiently large training set ${\cal D}_N$, generated by a bandlimited process $f(\mathbf{x})$, and the input $\mathbf{x}$ and all model parameters of the neural network are bounded. If the neural network $\tilde{f}(\mathbf{x})$ yields zero empirical risk on ${\cal D}_N$:
$$
R_{emp} (\tilde{f} \, |\,  \mathcal{D}_N) = 0,
$$
then it surely yields zero expected risk as $N \to \infty$:
$$
\lim_{N \to \infty} \;\; R(\tilde{f}\, |\,  \mathcal{D}_N)  \longrightarrow 0.
$$
\end{corollary}

\section{Equivalence of Perfect Learning}

\begin{theorem} (equivalence)
\label{theorem-equivalence}
Assume that the target function $f(\mathbf{x})$ is strictly or approximately 
bandlimited and any two bandlimited (either strictly or approximately) models, $\hat{f}_1(\mathbf{x})$ and $\hat{f}_2(\mathbf{x})$, are learned from a sufficiently large training set ${\cal D}_N$. If both models yield zero empirical risk on ${\cal D}_N$:
$$
R_{emp} (\hat{f}_1 \, |\,  \mathcal{D}_N) =  R_{emp} (\hat{f}_2 \, |\,  \mathcal{D}_N) = 0,
$$
then $\hat{f}_1(\mathbf{x})$ and $\hat{f}_2(\mathbf{x})$ are asymptotically identical under $p(\mathbf{x})$  as $N \to \infty$:
$$
\lim_{N \to \infty} \;\idotsint_{-\infty}^{+\infty}  \left\Vert \hat{f}_1(\mathbf{x}) - \hat{f}_2(\mathbf{x}) \right\Vert^2  p(\mathbf{x}) \; d\mathbf{x} \;\; 
\longrightarrow 0.
$$
\end{theorem}

{\bf Proof sketch:} 
According to the uniqueness theorem in mathematical analysis \cite{Levin64} ,  
as long as the sampled points are dense enough in the space, there exists a unique bandlimited function that may exactly pass through all of these samples. 
See the full proof in Appendix \ref{appendix-proof-equivalence}. 
$\hspace*{\fill} {\hfill\blacksquare}$

This result suggests that we may use many different models to solve a real-world machine learning problem.  As long as these models are powerful enough to act as a universal approximator to fit well to any given large training set, they are essentially equivalent as long as they reveal the bandlimiting behaviour, no matter whether you use a recurrent or nonrecurrent structure, use 50 layers or 100 layers in the model, etc. The key is how to apply the heavy engineering tricks to fine-tune the learning process to ensure that the complicated learned models fit well to the large training set. 

\section{Non-ideal Cases with Noises}

In this work, we mainly focus on the ideal learning scenarios where no noise is involved in the learning process. In practice, the collected training samples are inevitably corrupted by all sorts of noises. For example, both inputs, $\mathbf{x}$, and outputs, $y$, of the target function $y=f(\mathbf{x})$ may be corrupted by some independent noise sources. 
These noise sources may have wider or even unlimited band. Obviously, these independent noises will impair the learning process. However, the above perfect learning theory can be extended to deal with noises. 
These cases will be further explored as our future work. 

\section{Final Remarks}

In this paper, we have presented some theoretical results to explain the success of large-scale supervised learning. This success is largely attributed to the fact that these real-world tasks are not as hard as we originally thought because they all arise from real physical processes that are bounded by the bandlimiting property. Even though all bandlimited supervised learning problems in the real world are asymptotically solvable in theory, we may not afford to collect sufficient training data to solve some of them in near future if they have a very high level of difficulty as determined by the band limit and the data distribution. It is an interesting question on how to predict such difficulty measures for  real-world tasks. Another interesting problem is how to explicitly bandlimit the models during the learning process. This issue may be critical to achieve effective learning when the training set is not large enough to ensure the asymptotic generalization suggested in Theorem 3. We conjecture that all regularization tricks widely used in machine learning may be unified under the idea of bandlimiting models. 

%\clearpage
%\newpage 

\begin{center}
\Large \bf Appendix
\end{center}

\appendix

\section{Proof of Theorem 1 (existence)}
\label{appendix-proof-theorem1}

Here we give the full proof of {\bf Theorem 1} regarding the existence of perfect supervised learning.

{\it Proof:} 
First of all, since $ p(\mathbf{x})$ is a p.d.f. in $\mathbb{R}^K$, for any arbitrarily small number $\varepsilon >0$, it is always possible to find a {\it bounded} region in $\mathbb{R}^K$, denoted as $ \Omega$ ($\subset \mathbb{R}^K $), to ensure that the total probability mass outside $\Omega$ is smaller than $\varepsilon$:
$
\idotsint_{\mathbf{x} \notin \Omega} p(\mathbf{x}) \; d\mathbf{x} < \varepsilon.
$

Secondly, since $f(\mathbf{x})$ is bandlimited by a finite $B$, we may partition the entire space $\mathbb{R}^K$ into a equally-spaced criss-cross grid formed from all dimensions of $\mathbf{x}$. The grid is evenly separated by $\pi/B$ in each dimension. This uniform grid partitions the whole space, $\mathbb{R}^K$. According to high-dimensional sampling theorem \cite{Petersen62}, if we sample the function $f(\mathbf{x})$ at all mesh points in the grid, the entire function can be fully restored. Moreover, the non-uniform sampling results in \cite{Yen56,Marvasti01} allows us to fully restore the function not just from the exact samples at the mesh points but from any one point in a near neighbourhood around each mesh point. Each neighbourhood of a mesh point is named as a cell. 
These cells belong to two categories: i) $\boldsymbol{\Theta}_0 $ includes all cells intersecting with $\Omega$; ii) $\boldsymbol{\Theta}_1$ includes the other cells not intersecting with $\Omega$. Based on \cite{Yen56,Marvasti01},  assume we can pick up at least one data point, $\mathbf{x}_i$, from each cell ${\bf c}_i$, and use them as nodes to form the multivariate interpolation series
as follows:
\begin{eqnarray} \label{eq-language-series}
\tilde{\bf f}(\mathbf{x}) & =&  \sum_{i=1}^{\infty} \;\;f(\mathbf{x}_i)   \Phi_i(\mathbf{x})  \\
& = &  \underbrace{\sum_{{\bf c}_i \in \boldsymbol{\Theta}_0 } \;\;f(\mathbf{x}_i)  \Phi_i(\mathbf{x})}_{\hat{f}(\mathbf{x})}
+ \underbrace{\sum_{{\bf c}_i \in \boldsymbol{\Theta}_1 } \;\;f(\mathbf{x}_i)   \Phi_i(\mathbf{x})}_{\hat{g}(\mathbf{x})} \nonumber
\end{eqnarray}
where $\Phi_i(\mathbf{x})$ is the basic interpolation functions, such as the cardinal interpolation functions in \cite{Petersen62}, or the fundamental Lagrande polynomials \cite{Sauer95,Gasca00}. The choice of the interpolation function $\Phi_i(\mathbf{x})$ ensures that they satisfy 
the so-called Cauchy condition:
\begin{equation} \label{eq-Cauchy-condition}
\Phi_i(\mathbf{x}_k) = \left\{
     \begin{array}{lc}
       1 & i = k \\
       0 & i\neq k \\
     \end{array}
\right.
\end{equation}

Since $f(\mathbf{x})$ is bandlimited by $B$, namely an entire function with finite exponent type $B$, and each node $\mathbf{x}_i$ is chosen from one distinct cell, namely the set of all nodes is an R-set \cite{Levin64} (Chapter II, \textsection 1),  according to \cite{Petersen62} and \cite{Levin64}  (Chapter IV, \textsection 4) , the interpolation series in eq.(\ref{eq-language-series}) converges uniformly into $f(\mathbf{x})$:
$%\begin{equation}
f(\mathbf{x}) = \tilde{\bf f}(\mathbf{x}) = \hat{f}(\mathbf{x}) + \hat{g}(\mathbf{x}) 
$%\end{equation}

Next, instead of deterministically choosing one node per cell, let's consider the case where all the nodes $\mathbf{x}_i$ are randomly chosen from the given p.d.f. $p(\mathbf{x})$. Since the bounded region $\Omega$ is partitioned into many non-empty cells, the total number of cells in $\boldsymbol{\Theta}_0$ must be finite. Assume there are $M$ cells within $\boldsymbol{\Theta}_0$ in total, let's denote them as
$\{ \mathbf{c}_1, \mathbf{c}_2, \cdots,  \mathbf{c}_M \}$.
If we randomly draw one sample, the probability of having it from cell $\mathbf{c}_m$  ($1\leq m \leq M$) is computed as
$\epsilon_m = \int_{\mathbf{x} \in \mathbf{c}_m}  p(\mathbf{x})  d\mathbf{x} \neq 0$.
If we draw $N$  ($N >M$) independent samples, the probability of $k$ ($1\leq k\leq M$) cells remaining empty 
is computed as:
$L_k = \frac{1}{k!} \sum^{j_1=M,\cdots, j_k=M}_{j_1=1, \cdots, j_k=1, j_1 \neq j_2 \cdots \neq j_k} (1-\epsilon_{j_1}-\epsilon_{j_2} - \cdots - \epsilon_{j_k})^N$.
Thus,  based on the inclusion-exclusion principle, after $N$ samples, the probability of no cell being left empty may be computed as:
$
\Pr(\mathbf{no\ empty\ in\ } \boldsymbol{\Theta}_0)  =  1 - \sum_{k=1}^{M-1} (-1)^{k+1} L_k.
$
Because $M$ is finite and fixed, it is easy to show as $N \to \infty$,  we have $\Pr(\mathbf{no\ empty\ in\ } \boldsymbol{\Theta}_0) \to 1$. In other words, as $N \to \infty$, we will surely have at least one sample from each cell in $\boldsymbol{\Theta}_0$ to precisely construct  $\hat{f}(\mathbf{x})$ in eq.(\ref{eq-language-series}), which is guaranteed to occur in probability as $N \to \infty$. 
Now, let's construct the interpolation function only using $N$  points in $\boldsymbol{\Theta}_0$:
\begin{equation} \label{eq-Lagrange-random-N}
\hat{f}(\mathbf{x}) = \sum_{{\bf c}_i \in \boldsymbol{\Theta}_0 } \;\;f(\mathbf{x}_i)  \cdot \Phi_i(\mathbf{x}).
\end{equation}  

In the following, we will prove that $\hat{f}(\mathbf{x})$ constructed as such satisfy all requirements in {\bf Theorem 1}.

Firstly, since the interpolation functions $\Phi_i(\mathbf{x})$ satisfy the Cauchy condition in eq.(\ref{eq-Cauchy-condition}), thus,
it is straightforward to verify $R_{emp} (\hat{f} \, |\,  \mathcal{D}_N) = 0$.

Secondly, assume we have drawn $N$ samples from $\boldsymbol{\Theta}_0$, we will show the contribution of $\hat{g}(\mathbf{x})$ in eq.(\ref{eq-language-series}) tends to be negligible as $N \to \infty$. 
Based on the estimates of truncation errors in sampling in \cite{Long2004,Brown69},
we have 
$$
\Vert f(\mathbf{x})  - \hat{f}(\mathbf{x} )  \Vert \leq O(N_0^{-\delta}) 
$$
where $\delta>0$ and $N_0$ denotes the minimum number of different projections of $N$ samples across any $K$ orthogonal axes in $\mathbb{R}^K$. Since all $N$ data samples are randomly selected, as $N \to \infty$, we are sure $N_0 \overset{P}{\longrightarrow} \infty$.
Putting all of these together, $\forall \mathbf{x} \in \boldsymbol{\Theta}_0 $, we have 
$$
\Vert \hat{f}(\mathbf{x} ) - f(\mathbf{x}) \Vert  \leq O(N_0^{-\delta})  \; \overset{P}{\longrightarrow}  0   
$$
as $N \to \infty$.

Finally, the expected risk of $\hat{f}(\mathbf{x})$ is calculated as:
\begin{eqnarray} \label{eq-R-conbergence}
R(\hat{f}) & =  &  \idotsint_{\mathbf{x} \in \boldsymbol{\Theta}_0} 
\Vert f(\mathbf{x}) - \hat{f}(\mathbf{x}) \Vert^2
p(\mathbf{x}) \; d\mathbf{x} \nonumber \\
& + & 
\idotsint_{\mathbf{x} \in \boldsymbol{\Theta}_1} 
\Vert f(\mathbf{x}) - \hat{f}(\mathbf{x})  \Vert^2
p(\mathbf{x}) \; d\mathbf{x} \nonumber \\
& \leq & O(N_0^{-\delta}) + 4 H^2 \cdot \epsilon 
\end{eqnarray}
where $H$ denotes the maximum value of the target function, i.e., $H = \sup_{\mathbf{x}} |f(\mathbf{x})|$.

Because $\epsilon$ in the second term may be made to be arbitrarily small in the first step when we choose $\Omega$,  we have 
$$
\lim_{N \to \infty}  R(\hat{f}) \overset{P}{\longrightarrow} 0.
$$

Therefore, we have proved the Langrage interpolation $\hat{f}(\mathbf{x})$ in eq.(\ref{eq-Lagrange-random-N}) using the randomly sampled $N$ points, ${\cal D}_N$,
satisfy all requirements in {\bf Theorem 1}.

If $f(\mathbf{x})$ is approximately bandlimited, the above proof also holds. The only change is to choose an approximate band limit $B_\epsilon$ to ensure the out-of-band probability mass $\epsilon$ is arbitrarily small. Then we just use $B_\epsilon$ to partition $\Omega$ in place of $B$. Everything else in the proof remains valid.  
$\hspace*{\fill} {\hfill\blacksquare}$

\section{Proof of Theorem 2}
\label{appendix_derivatives_eq}

{\it Proof:} 
If the target function $y=f(\mathbf{x})$ is strictly bandlimited, it must be an analytic function in $\mathbb{R}^K$. We may expand $y=f(\mathbf{x})$ as the Taylor series according to Taylor's Theorem in several variables. For notation simplicity, we adopt the well-known multi-index notation \cite{Sauer95} to represent the exponents of several variables. A multi-index notation is a $K$-tuple of nonnegative integers, denoted by a Greek letter such as $\alpha$: $\alpha = ( \alpha_1, \alpha_2, \cdots, \alpha_K)$ with $\alpha_k \in \{0,1,2,\cdots \}$. If $\alpha$ is a multi-index, we define 
$|\alpha| = \alpha_1 + \alpha_2 + \cdots + \alpha_K$,
$\alpha ! = \alpha_1 ! \alpha_2 ! \cdots \alpha_K!$,
$\mathbf{x}^\alpha = x_1^{\alpha_1} x_2^{\alpha_2} \cdots x_K^{\alpha_K}$ (where $\mathbf{x} = (x_1 x_2 \cdots x_K) \in \mathbb{R}^K$), and
$\partial^\alpha f(\mathbf{x})
=\partial_1^{\alpha_1} \partial_2^{\alpha_2} \cdots \partial_K^{\alpha_K} f(\mathbf{x})
=\frac{\partial^{|\alpha|} f(\mathbf{x})}{\partial x_1^{\alpha_1} x_2^{\alpha_2} \cdots x_K^{\alpha_K} }$. The number $|\alpha|$ is called the order of $\alpha$.

According to Taylor's Theorem in several variables, $y=f(\mathbf{x})$ may be expanded around any point $\mathbf{x}_0 \in \mathbb{R}^K$ as follows:
\begin{equation}
f(\mathbf{x}) = \sum_{|\alpha| = 0}^{\infty} \;\;
\frac{\partial^\alpha f(\mathbf{x}_0)}{\alpha !} \; (\mathbf{x} -\mathbf{x}_0 )^\alpha.
\end{equation}
Because the function $y=f(\mathbf{x})$ is bandlimited by $B$, according to the Bernstein's inequality on Page 138 of \cite{Achiester56},
we know the coefficients in the above Taylor series satisfy:
\begin{equation} \label{eq-bandlimiting-condition-polynomial}
\Vert \partial^\alpha f(\mathbf{x}_0) \Vert \leq B^{|\alpha|}  \cdot H
\end{equation}
for all $\alpha$ from $|\alpha|=0,1,2, ....$, and $H = \sup_{\mathbf{x}} |f(\mathbf{x})|$.

Given the $N$ samples, ${\cal D}_N = \{\mathbf{x}_1, \mathbf{x}_2, \cdots, \mathbf{x}_N \}$, ($ \mathbf{x}_i \in \mathbb{R}^K $), assume we may have an ideal learning algorithm to construct a new model  $\hat{f}(\mathbf{x} | {\cal D}_N)$. The optimal function $\hat{f}_n^*(\mathbf{x} | {\cal D}_N)$ should be the Taylor polynomial of $f(\mathbf{x})$ with the order of $n$. We assume the problem is poised with respect to the given ${\cal D}_N$ \cite{Sauer95,Gasca00} \footnote{If all data points in ${\cal D}_N$ are randomly sampled, the problem is poised in probability 1.}, we need to have the same number of (or slightly more) free coefficients in polynomials as the total number of data points in ${\cal D}_N$, namely ${n+K \choose K} = N$, we may compute $n$ roughly as $n \simeq O(N^\frac{1}{K})$. In other words, the optimal model may be represented as a multivariate polynomial:
\begin{equation} \label{eq-model-as-Taylor-polynomial}
\hat{f}_n^*(\mathbf{x} | {\cal D}_N) = \sum_{|\alpha| = 0}^n \;\;
c_{\alpha}\; (\mathbf{x} -\mathbf{x}_0 )^\alpha
\end{equation}
where each coefficient $c_{\alpha} = \frac{\partial^\alpha f(\mathbf{x}_0)}{\alpha !}$ for all $\alpha$ up to the order of $n$. As in \cite{Sauer95}, if the problem is poised with respect to ${\cal D}_N$, these Taylor polynomial coefficients may be uniquely determined by the $N$ training samples in ${\cal D}_N$.

As a side note, we may see why bandlimitedness and model complexity are two different concepts. The model complexity is determined by the number of free model parameters. When representing a model as a multivariate Taylor polynomial in eq.(\ref{eq-model-as-Taylor-polynomial}), the model complexity is determined by the total number of free coefficients, $c_{\alpha}$, in the expansion. The higher order $n$ we use, the more complex model we may end up with. However, no matter what order $n$ is used, as long as all coefficients satisfy the contraints in eq.(\ref{eq-bandlimiting-condition-polynomial}) and other constraints in \cite{Veron1994}, the resultant model is bandlimited by $B$.

Based on the remainder error in the multivariate Taylor's theorem, we have 
\begin{equation} \label{eq-taylor-remainder-error}
f(\mathbf{x}) - \hat{f}_n^*(\mathbf{x})  = 
\sum_{|\alpha|=n+1} 
\frac{\partial^\alpha f(\mathbf{x}_0 + \xi \cdot \mathbf{x} )}{\alpha !} \; (\mathbf{x} -\mathbf{x}_0 )^\alpha
\end{equation}
for some $\xi \in (0,1)$. Since $f(\mathbf{x})$ is bandlimited by $B$ and $|\alpha|=n+1$, we have 
$\Vert \partial^\alpha f(\mathbf{x}_0 + \xi \cdot \mathbf{x} ) \Vert
\leq B^{n+1} \cdot H
$. 
Furthermore, since $\mathbf{x} \sim {\cal N} (0, \sigma^2 {\bf I})$, we choose $\mathbf{x}_0 =0$, 
and after applying the multinomial theorem,  we have
\begin{equation} \label{eq-taylor-remainder-error-bound}
\Vert f(\mathbf{x}) - \hat{f}_n^*(\mathbf{x}) \Vert
\leq \frac{B^{n+1} \cdot H }{ (n+1)!}
||\mathbf{x}||^{n+1}.
\end{equation}
where $||\mathbf{x}|| = |x_1| + |x_2| + \cdots + |x_K|$. 

Then the perfect learning algorithm yields:

\begin{eqnarray} 
{\cal R}^*_N & \leq  & {\bf E}_{ \mathcal{D}_N } \left[   \mathbf{E}_{\mathbf{x}} \left[ \Vert\hat{f}(\mathbf{x} | \mathcal{D}_N) - f^*_n(\mathbf{x}) \Vert ^2  \right]     \right] \nonumber \\
& = & \mathbf{E}_{\mathbf{x}} \left[
\left( \frac{B^{n+1} \cdot H }{ (n+1)!} \right)^2
||\mathbf{x}||^{2n+2} \right] \nonumber \\
& \leq &  \left(\frac{B^{n+1} \cdot H }{ (n+1)!} \right)^2 K^{2n+2} \mathbf{E} 
\left( |x_k|^{2n+2}  \right)  (Radon's\ ineq) \nonumber \\
& = & \left(\frac{(KB)^{n+1}\cdot H }{ (n+1)!} \right)^2 \cdot
\sigma^{2n+2} \cdot 2^{n+1} \frac{\Gamma(n+1+\frac{1}{2})}{\sqrt{\pi}} \nonumber \\
& = & 
\frac{(2 K^2 B^2\sigma^2)^{n+1} \cdot H^2}
{(n+1)!}
{{n+\frac{1}{2}} \choose n+1}  \nonumber \\
& < &
\left[ \frac{(\sqrt{2} K B\sigma)^{n+1} \cdot H}
{\sqrt{(n+1)! }}
 \right] ^2 \nonumber
\end{eqnarray}

Refer to \cite{Winkelbauer12} for the central absolute moments of normal distributions, $\mathbf{E} 
( |x_k|^{2n+2})$.
$\hspace*{\fill} {\hfill\blacksquare}$

\section{Proof of Theorem 3 (sufficient condition)}
\label{appendix-proof-condition}

{\it Proof:} We first consider the strictly bandlimited case: assume the target function $f(\mathbf{x})$ is bandlimited by $B$ and the learned model $\hat{f}(\mathbf{x})$ is bandlimited by $B'$. According to eq.(\ref{eq-model-as-Taylor-polynomial}), the strictly bandlimited function $f(\mathbf{x})$ can be expanded around any $\mathbf{x}_0$ as the Taylor's series:
$$
f(\mathbf{x}) = \sum_{|\alpha| = 0}^{\infty}  b_{\alpha}\; (\mathbf{x} -\mathbf{x}_0 )^\alpha
$$
where $b_{\alpha}=\frac{\partial^\alpha f(\mathbf{x}_0)}{\alpha !}$ for all $\alpha$ up to $\infty$. Since $f(\mathbf{x})$ is bandlimited by $B$, we have 
$\Vert \partial^\alpha f(\mathbf{x}_0) \Vert
\leq B^{|\alpha|} \cdot H$. Obviously, we have 
$|b_{\alpha}| \leq \frac{B^{|\alpha|}\cdot H}{\alpha !} \to 0 $ as $|\alpha| \to \infty$.  Therefore, a bandlimited function may be represented as an infinite sum of orthogonal base functions. Since the coefficients of these terms are decaying, the series may be truncated and approximated by a finite partial sum of $n$ terms up to arbitrary precision (as $n$ goes large). 
\begin{equation} \label{eq-target-function-taylor}
f(\mathbf{x}) = \sum_{|\alpha| = 0}^{n}  b_{\alpha}\; (\mathbf{x} -\mathbf{x}_0 )^\alpha
+ \xi_n(\mathbf{x})
\end{equation}
where $\xi_n(\mathbf{x})$ denotes the remainder term in the Taylor's expansion. 

Let's assume the input $\mathbf{x} \sim p(\mathbf{x})$ is constrained in a bounded region in $\mathbb{R}^K$, we may normalize all $\mathbf{x}$ within a hypercube $[-U,U]^K$. Similar to the remainder error in eq.(\ref{eq-taylor-remainder-error-bound}), we can easily derive:
$$
\Vert \xi_n(\mathbf{x}) \Vert \leq \frac{(K B U)^{n+1} H}{(n+1)!} \to 0
$$
as $n \to \infty$. 

Similarly, since the learned model $\hat{f}(\mathbf{x})$ is also bandlimited by $B'$, we may expand it in the same way as:
\begin{equation} \label{eq-learned-model-taylor}
\hat{f}(\mathbf{x}) = \sum_{|\alpha| = 0}^{n}  d_{\alpha}\; (\mathbf{x} -\mathbf{x}_0 )^\alpha
+ \xi'_n(\mathbf{x}).
\end{equation}
where we have $ \lim_{|\alpha| \to \infty} |d_{\alpha}| \to 0 $, and $\Vert \xi'_n(\mathbf{x}) \Vert \leq \frac{(K B' U)^{n+1} H}{(n+1)!} \to 0$ as $n \to \infty$. 

Given ${\cal D}_N = \{ (\mathbf{x}_1,y_1),  (\mathbf{x}_2,y_2), \cdots, (\mathbf{x}_N,y_N)\} $, a training set of $N$ samples,
$n$ may be chosen as such, $n \simeq O(N^\frac{1}{K})$ , to have exactly $N$ terms in the partial sums in eqs. (\ref{eq-target-function-taylor}) and (\ref{eq-learned-model-taylor}). Since all training samples are generated by the target function $f(\mathbf{x})$, thus we have:
$$
y_j = f(\mathbf{x}_j) = \sum_{|\alpha| = 0}^{n}  b_{\alpha}\; (\mathbf{x}_j -\mathbf{x}_0)^\alpha
+ \xi_n(\mathbf{x}_j) \; (j=1,\cdots,N).
$$

Meanwhile, if the model $\hat{f}(\mathbf{x})$ is learned to yield zero empirical loss in ${\cal D}_N$, then $\hat{f}(\mathbf{x})$ also fits to every sample in ${\cal D}_N$ as follows:
$$
y_j = \hat{f}(\mathbf{x}_j) = \sum_{|\alpha| = 0}^{n}  d_{\alpha}\; (\mathbf{x}_j -\mathbf{x}_0)^\alpha
+ \xi'_n(\mathbf{x}_j) \; (j=1,\cdots,N).
$$

Taking difference between each pair of them, we may represent the results as the following matrix format:
$$
  \begin{bmatrix}  (\mathbf{x}_1 -\mathbf{x}_0)^{\alpha_1}  & \cdots &  (\mathbf{x}_1 -\mathbf{x}_0)^{\alpha_N} \\ 
 	              & \vdots &   \\
 			      (\mathbf{x}_j -\mathbf{x}_0)^{\alpha_1}  & \cdots &  (\mathbf{x}_j -\mathbf{x}_0)^{\alpha_N} \\
 				  & \vdots &  \\
 				   (\mathbf{x}_N -\mathbf{x}_0)^{\alpha_1}  & \cdots &  (\mathbf{x}_N -\mathbf{x}_0)^{\alpha_N} \\
 \end{bmatrix}_{N \times N}
  \begin{bmatrix}  b_{\alpha_1} - d_{\alpha_1} \\ 
 	              \vdots  \\
 			       b_{\alpha_j} - d_{\alpha_j} \\
 				  \vdots  \\
 				   b_{\alpha_N} - d_{\alpha_N} \\
 \end{bmatrix}_{N \times 1}
$$
$$
 =
  \begin{bmatrix}  \xi'_n(\mathbf{x}_1)  - \xi_n(\mathbf{x}_1) \\ 
 	              \vdots  \\
 			       \xi'_n(\mathbf{x}_j)  - \xi_n(\mathbf{x}_j) \\
 				  \vdots  \\
 				   \xi'_n(\mathbf{x}_N)  - \xi_n(\mathbf{x}_N) \\
 \end{bmatrix}_{N \times 1} 
 = \boldsymbol{\xi}_N
$$ 

The $ N\times N$ matrix in the left-hand side is the so-called multivariate Vandermonde matrix where all column  vectors are constructed from orthogonal multivariate Taylor base functions.  When all $\mathbf{x}_j$  in ${\cal D}_N$
are randomly drawn from $p(\mathbf{x})$,  
as in \cite{Sauer95}, the problem is poised with respect to ${\cal D}_N$ in probability one. Thus, this $ N\times N$ matrix has full rank and is invertible. Meanwhile, as $N \to \infty$, the $N \times 1$ vector in the right hand side approaches 0, i.e. $ \lim_{N \to \infty}\boldsymbol{\xi}_N = 0$.  Therefore, we may deduce that all coefficients converge as $d_{\alpha_j} = b_{\alpha_j}$ for all $j=1,2, \cdots, N$ as $N \to \infty$. In other words, 
the learned model $\hat{f}(\mathbf{x})$ converges towards the target function $f(\mathbf{x})$ except those negligible high-order terms. As a result, we can show the expected loss as:
\begin{eqnarray}
R(\hat{f}\, |\,  \mathcal{D}_N)   & = & \idotsint_{-\infty}^{+\infty}
\left\Vert f(\mathbf{x}) - \hat{f}(\mathbf{x}) \right\Vert^2 
p(\mathbf{x}) \; d\mathbf{x} \nonumber \\
& \leq & \left[ \frac{(K B U)^{n+1} H + (K B' U)^{n+1} H  }{(n+1)!} \right]^2 \to 0 \nonumber 
\end{eqnarray}
as $N \to \infty$.

If either $f(\mathbf{x})$ or $\hat{f}(\mathbf{x})$ is approximately bandlimited, the above proof also holds. The only change is to choose an approximate band limit $B_\epsilon$ to ensure the out-of-band residual $\epsilon$ is 
arbitrarily small. Then, we just use $B_\epsilon$ or $B'_\epsilon$ in place of $B_\epsilon$ or $B'_\epsilon$. Therefore, we conclude that $\lim_{N \to \infty} R(\hat{f} | \mathcal{D}_N) =0$ holds for either strictly or approximately bandlimited target functions and learned models.
$\hspace*{\fill} {\hfill\blacksquare}$

\section{Proof of Theorem 4 (equivalence)}
\label{appendix-proof-equivalence}
{\it Proof:} 
Based on Theorem 3, we have 
$$
\lim_{N \to \infty}  R(\hat{f}_1) \longrightarrow 0
$$
and 
$$
\lim_{N \to \infty}  R(\hat{f}_2) \longrightarrow 0.
$$

Therefore, we have
\begin{eqnarray}
&& \lim_{N \to \infty} \;\idotsint_{-\infty}^{+\infty}  \left\Vert \hat{f}_1(\mathbf{x}) - \hat{f}_2(\mathbf{x}) \right\Vert^2  p(\mathbf{x}) \; d\mathbf{x}  \nonumber \\
& \leq &
 \lim_{N \to \infty} \;\idotsint_{-\infty}^{+\infty}  \left\Vert \hat{f}_1(\mathbf{x}) - f(\mathbf{x}) \right\Vert^2  p(\mathbf{x}) \; d\mathbf{x} \nonumber \\
 && + 
 \lim_{N \to \infty} \;\idotsint_{-\infty}^{+\infty}  \left\Vert \hat{f}_2(\mathbf{x}) - f(\mathbf{x}) \right\Vert^2  p(\mathbf{x}) \; d\mathbf{x} \nonumber \\
& =  &\lim_{N \to \infty}  R(\hat{f}_1) + \lim_{N \to \infty}  R(\hat{f}_2)  
 \;\;\;  \overset{P}{\longrightarrow} 0. 
\end{eqnarray}  
$\hspace*{\fill} {\hfill\blacksquare}$

%\section*{Acknowledgments} 
% 
%\textbf{Do not} include acknowledgements in the initial version of
%the paper submitted for blind review.
%
%If a paper is accepted, the final camera-ready version can (and
%probably should) include acknowledgements. In this case, please
%place such acknowledgements in an unnumbered section at the
%end of the paper. Typically, this will include thanks to reviewers
%who gave useful comments, to colleagues who contributed to the ideas, 
%and to funding agencies and corporate sponsors that provided financial 
%support.  

\nocite{Davis63,Micchelli79,Klamer79,Leneman66}

\bibliography{LimitedBand}
\bibliographystyle{icml2015}

\end{document}